\pdfoutput=1
\documentclass[11pt]{article}

\usepackage[]{acl}

\usepackage{times}
\usepackage{latexsym}

\usepackage[T1]{fontenc}
\usepackage[utf8]{inputenc}

\usepackage{microtype}
\usepackage{graphicx}
\usepackage{amsmath}
\usepackage{amssymb}
\usepackage{booktabs}
\usepackage{xcolor}
\usepackage{multirow}
\usepackage{balance}
\usepackage{pifont}
\usepackage[linesnumbered,lined,ruled,noend]{algorithm2e}
\usepackage[linesnumbered]{algorithm2e}
\makeatletter
\newcommand{\nosemic}{\renewcommand{\@endalgocfline}{\relax}}% Drop semi-colon ;
\newcommand{\dosemic}{\renewcommand{\@endalgocfline}{\algocf@endline}}% Reinstate semi-colon ;
\let\oldnl\nl% Store \nl in \oldnl
\newcommand{\nonl}{\renewcommand{\nl}{\let\nl\oldnl}}% Remove line number for one line
\makeatother
\title{LLM-based Frameworks for API Argument Filling \\ in Task-Oriented Conversational Systems}
\author{Jisoo Mok$^1$\thanks{~~Work done while interning at Amazon (magicshop1118@snu.ac.kr)} ~~~~~~~ Mohammad Kachuee$^{2}$ ~~~~~~~ Shuyang Dai$^2$ ~~~~~~~ Shayan Ray$^2$ \\ ~~~~~~ \textbf{Tara Taghavi}$^{2}$  ~~~~~~~ \textbf{Sungroh Yoon}$^{1,4}$\thanks{~~Corresponding Authors}\\
$^1$ Department of ECE, Seoul National University ~~~ $^2$ Amazon \\
$^4$ Interdisciplinary Program in Artificial Intelligence, Seoul National University}

\begin{document}
\maketitle
\begin{abstract}
% Task-oriented conversational agents consist of three main components: external API selection, argument filling, and response generation. 
Task-orientated conversational agents interact with users and assist them via leveraging external APIs. A typical task-oriented conversational system can be broken down into three phases: external API selection, argument filling, and response generation. The focus of our work is the task of argument filling, which is in charge of accurately providing arguments required by the selected API. Upon comprehending the dialogue history and the pre-defined API schema, the argument filling task is expected to provide the external API with the necessary information to generate a desirable agent action. In this paper, we study the application of Large Language Models (LLMs) for the problem of API argument filling task. Our initial investigation reveals that LLMs require an additional grounding process to successfully perform argument filling, inspiring us to design training and prompting frameworks to ground their responses. Our experimental results demonstrate that when paired with proposed techniques, the argument filling performance of LLMs noticeably improves, paving a new way toward building an automated argument filling framework.
\end{abstract}

\section{Introduction}
\begin{figure*}[t]
\begin{center}
  \includegraphics[width=\linewidth]{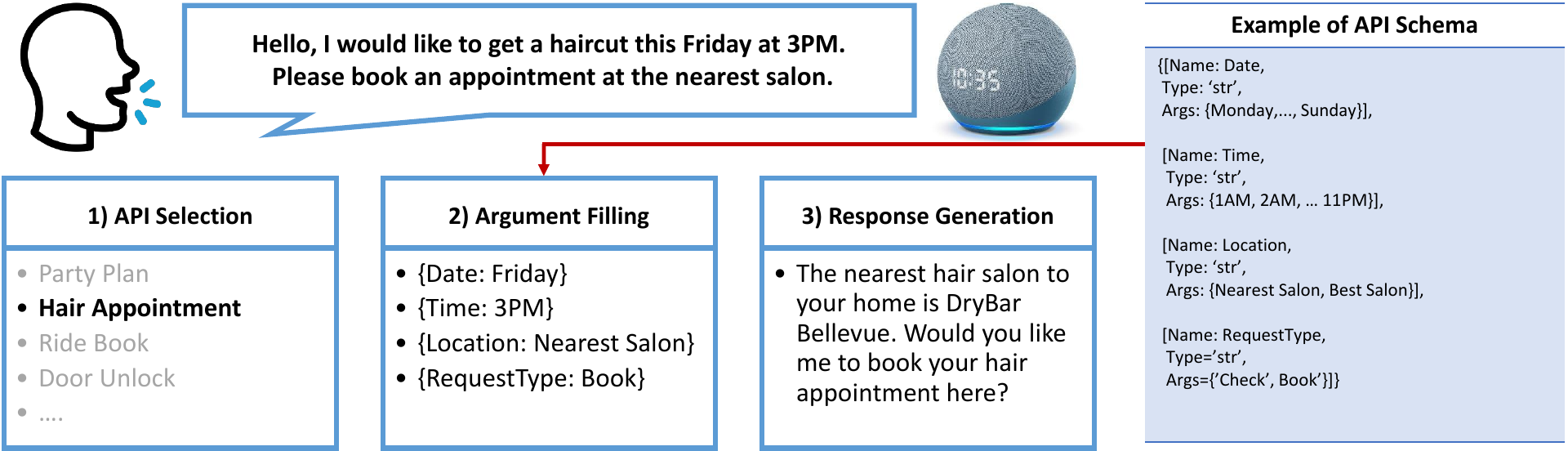}
\end{center}
\vspace{-10pt}
  \caption{An overview of how a task-oriented conversational agent might complete a user's request to book a haircut. To begin with, the agent selects the "Hair Appointment" API from the list of available APIs. An example of the pre-defined API schema associated with the "Hair Appointment" API is given on the far right side. Following API selection, the argument filling step utilizes the API schema and dialogue history to identify arguments to complete the API call. Finally, the agent responds to the user with the utterances produced in the response generation step.}
\vspace{-10pt}
\label{fig:argfilling_def}
\end{figure*}

Task-oriented conversational systems, illustrated in~\figurename~\ref{fig:argfilling_def}, largely consist of three processes: external API selection, argument filling, and response generation~\cite{hosseini2020simple}. 
The API selection phase selects which one from the pre-defined pool of APIs must be called to complete the user request.
Once the appropriate external API to carry out the user request has been selected, the argument filling phase must reliably identify and provide correct arguments to the API by faithfully following the API schema and dialogue history. 
An API schema, an example of which is also demonstrated in~\figurename~\ref{fig:argfilling_def}, is typically assumed to be given as a part of the API and includes required arguments and their types.
Therefore, the API schema and dialogue history provide sufficient information for the conversational agent to identify which arguments are necessary to complete the API call.
Lastly, the response generation phase, as the name suggests, returns an appropriate response to the user based on the API output.

\begin{figure}[t]
\begin{center}
  \includegraphics[width=\linewidth]{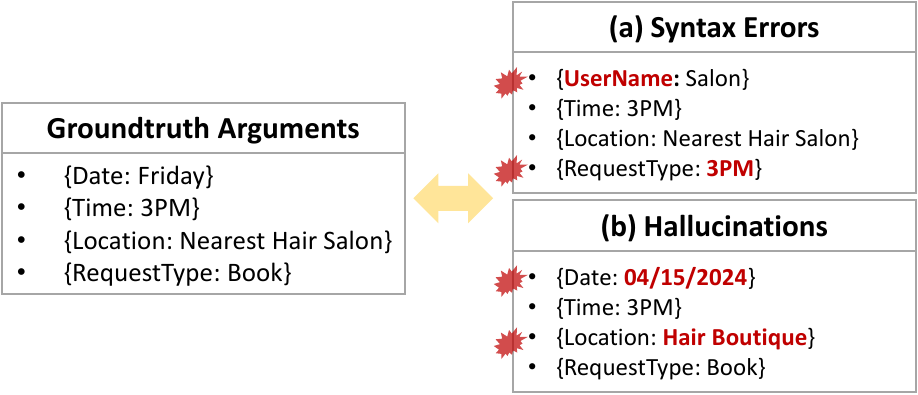}
\end{center}
\vspace{-10pt}
  \caption{Examples of two potential errors that can arise in argument filling. (a) Syntax errors refer to those that digress away from the pre-defined API schema. (b) Hallucinations correspond to those that deviate from the user intention and utterances.}
\vspace{-10pt}
\label{fig:error_types}
\end{figure}

The user dissatisfaction in argument filling mainly stems from the conversational agent being incapable of adhering to the API schema and dialogue history. 
The erroneous arguments that digress away from the API schema are considered "Syntax Errors", and hallucinated responses that deviate from the user utterances are considered "Hallucinations."
In~\figurename~\ref{fig:error_types}, we provide examples of each error type that occurs when performing argument filling for the "Hair Appointment" API.
% The required arguments are provided as a part of the API schema, 
% The argument filling phase is expected to fill in these pre-defined arguments with appropriate values following the API schema and user utterances.

Large Language Models (LLMs) trained with instructions have recently been garnering much attention as a promising model for enabling human-like and safe user-agent interactions in open-domain conversations~\cite{ouyang2022training, wang2022super}. 
The aim of this paper is to explore whether the strength of LLMs can be harnessed specifically for the purpose of argument filling in task-oriented conversational systems.
To construct an LLM-backed framework for argument filling, their outputs must strictly follow and stay faithful to the pre-defined API schema and user utterances, a process commonly known as ``grounding."
Our initial zero-shot performance evaluation of LLMs of various sizes reveals that LLM-generated responses suffer severely from both syntax errors and hallucinations, necessitating the development of additional techniques to appropriately ground their responses for the task of our interest.
% fail to generate appropriate responses 
% in which the conversational agent must be able to closely follow the pre-defined API schema and previous dialogue history when generating a response.
% Enforcing LLMs to strictly follow the pre-defined schema and user utterances involves a 
% their outputs on pre-defined schema and user utterances, which served as an inspiration for the title of our project: schema and user grounded argument filling.
% In this project, we first evaluated the zero-shot performance of LLMs of various sizes to provide a baseline performance for reference and explored an advanced prompting technique to study the effect of prompt designs on performance metrics. 

We investigate two separate and unique avenues to tackle the problem of grounding for open- and closed-sourced LLMs.
On one hand, for open-sourced LLMs,~\textit{e.g.,} LLAMA-v1-7B, we propose a two-step instruction-tuning framework that is comprised of supervised fine-tuning (SFT) and rejection sampling (RS).
Our experimental results show that utilizing the proposed instruction-tuning framework noticeably outperforms the na\"ive SFT baseline.
On the other hand, in the case of closed-sourced LLMs whose weights are not directly accessible, we demonstrate that their performance can be improved by replacing the plain prompt design with a "multi-step prompting" scheme.
% Afterwards, we fine-tuned the LLAMA-7B model, an open-source and relatively light-weight model, on task-oriented dialogue datasets to study how much supervised fine-tuning (SFT) can improve upon the baseline performance measures. Lastly, based on the insights from the above experiments, we designed a novel training framework that can avoid model overfitting and make use of model-generated responses, instead of relying on additional data, to further improve the performance metrics from the SFT baseline.
Our contributions can be summarized as follows:
\begin{itemize}
\item{This is the first work to explore the utilization of LLMs for argument filling in task-oriented conversational agents. Our results demonstrate that when paired with a proper grounding process, LLMs can offer a simpler and more autonomous alternative to conventional approaches in argument filling.}
\item{For open-sourced LLMs, we propose a cohesive training pipeline to ground their behaviors. The proposed training pipeline consists of two phases: model bootstrapping via supervised fine-tuning and additional fine-tuning with model-generated outputs, which have undergone rejection sampling through a custom reward function. For closed-sourced LLMs, we explore an advanced prompting technique that is more fine-grained and informative.}
\item{We provide substantial experimental results to demonstrate the effectiveness of the proposed approaches. Notably, the LLAMA-v1-7B model fine-tuned using the proposed instruction-tuning pipeline outperforms strong zero-shot baselines obtained by prompting significantly larger LLMs.}
% Further deep-dive analyses reveal that the proposed approach appropriately addresses the problem of schema- and user-grounded argument filling.}
\end{itemize}

\section{Related Works}
\subsection{Language Models for Task-oriented Dialogues}
Utilization of pre-trained Language Models for Task-oriented Dialogues (ToD) was pioneered by~\citet{zhang2019dialogpt} and~\citet{peng2021soloist}.
~\citet{kulhanek2021augpt} and~\citet{lin2020mintl} improved the basic ToD modeling approaches by employing contrastive state training and belief state differences, respectively.
Other works~\cite{pandey2018exemplar, cai2019retrieval, nekvinda2022aargh} proposed to combine generative models with retrieval-based approaches. 
While~\citet{hudevcek2023llms} perform zero-shot evaluation of various LLMs for ToD modeling, to the best of our knowledge, this is the first work to exploit and instruction-tune LLMs in the billion-parameter regime for argument filling in ToD systems.
% Several improvements to the basic setup were proposed, such as contrastive state training (Kulhánek
% et al., 2021) or using belief state differences (Lin
% et al., 2020). Others proposed a combination of
% generative models with retrieval-based approaches
% (Pandey et al., 2018; Cai et al., 2019; Nekvinda and
% Dušek, 2022). All described works finetune LMs
% on in-domain data, which is in contrast with the
% pure in-context learning approach that we apply

\begin{figure*}[t]
\begin{center}
  \includegraphics[width=\linewidth]{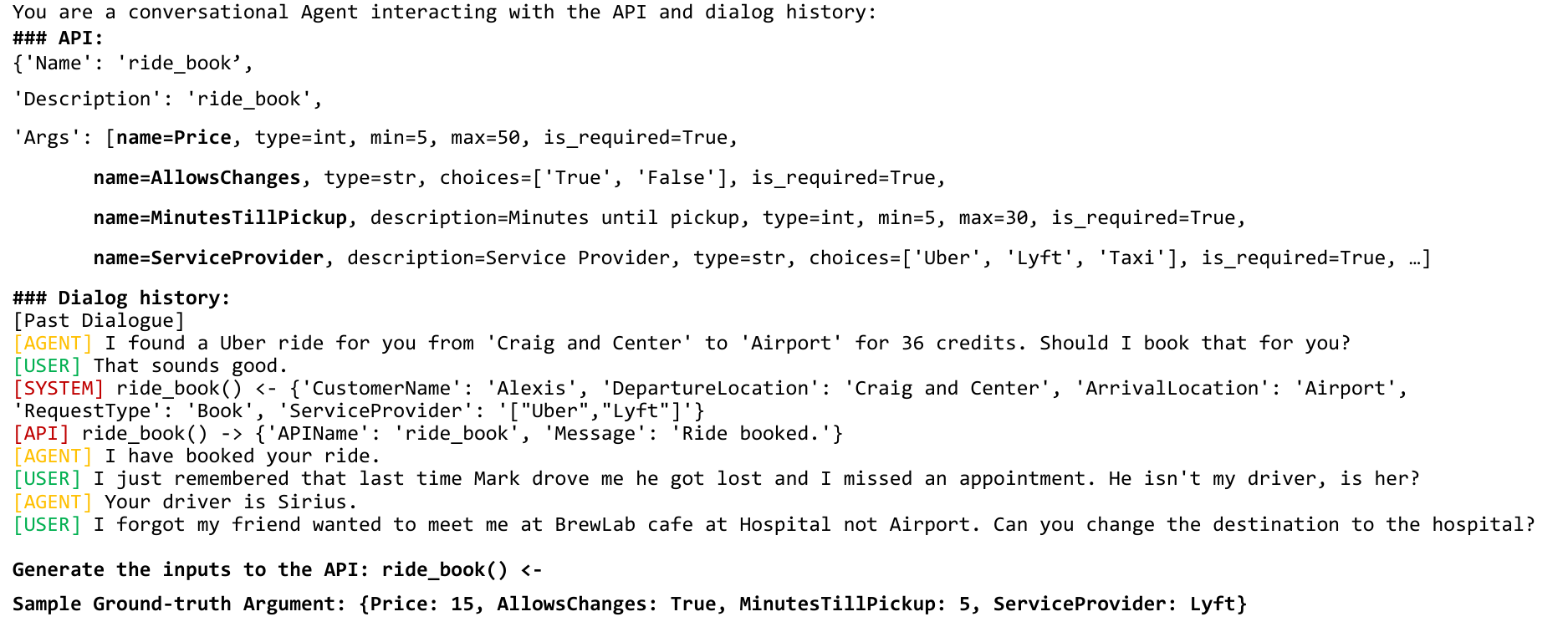}
\end{center}
\vspace{-10pt}
  \caption{Abbreviated illustration of the default prompt template that includes API description and dialogue history. We also provide an example of a ground-truth argument, which is pre-processed to follow a dictionary-like format.}
\vspace{-10pt}
\label{fig:prompt template}
\end{figure*}

\subsection{Large Language Models and Instruction-tuning}
The introduction of Transformer-based architectures heralded the beginning of large and incredibly capable models for Natural Language Processing (NLP)~\cite{vaswani2017attention}.
Transformer-based language models with several billions of parameters, such as GPT-3~\cite{brown2020language} and OPT~\cite{zhang2022opt}, have shown unprecedented zero- and few-shot performance across diverse NLP tasks.
The generalization capability of these so-called Large Language Models (LLMs) was further improved by training them via instruction-tuning~\cite{goldwasser2014learning} with in-context instructions.
The promising results obtained by instruction-tuning inspired the development of large instruction-paired datasets, such as NaturalInstructions-v1 (NI-v1)~\cite{NIv1} and SuperNaturalInstructions~\cite{NIv2}.
The remarkable performance of general-purpose instruction-tuned models inspired the development of more domain-specific models.
Examples of such models include: InstructUIE~\cite{wang2023instructuie} for information extraction, CoEDIT~\cite{raheja2023coedit}
for writing, ChatDoctor~\cite{yunxiang2023chatdoctor} for medical purposes, and Goat~\cite{liu2023goat} for mathematics.
% •	
% •	We propose a cohesive training pipeline to ground the LLMs’ behaviors on pre-defined schema and dialogue history. The proposed training pipeline consists of two phases: model strapping via supervised fine-tuning and additional fine-tuning with a subset of model-generated outputs, which are rejection-sampled through a custom reward function. 
% o	In the initial bootstrapping phase, we employ several techniques to circumvent model overfitting.
% o	We define a custom reward function to score model-generated outputs for rejection sampling. Only a subset of model-generated outputs that yield positive reward are used in the additional fine-tuning phase.
% •	We provide substantial experimental results to demonstrate the effectiveness of the proposed approach in improving argument filling performance under in- and out-of-domain benchmarks. Notably, the LLAMA-v1-7B model fine-tuned using the proposed pipeline outperforms strong baselines obtained by prompting significantly larger LLMs (e.g., LLAMA-v2-13B and ChatGPT). Further deep-dive analyses reveal that the proposed approach appropriately address the problem of schema- and user-grounded argument filling.
% \section{Related Works}
% \subsection{Task-oriented Dialogues}
% \subsection{LLMs and Their Applications}

\section{Proposed Methodology}
\subsection{Prompt Design}
To guarantee experimental consistency across different models and datasets, we first design a common prompt template for argument filling.
An example of the default prompt template, which includes a short instruction, the pre-defined API schema, and dialogue history up to the specified API call, is provided in~\figurename~\ref{fig:prompt template}.
This prompt template is used for both in-context instruction tuning and evaluation processes and remains fixed across all of our experiments unless stated otherwise. 
% and in-context instruction tuning and evaluation and fix it for all of our experiments. 
% An abbreviated example of our prompt design is provided in the figure below:

\subsection{Instruction-tuning Framework for Open-sourced LLMs}~\label{inst_tuning}
\noindent\textbf{Phase I. Model Bootstrapping via Supervised Fine-tuning} 
We first bootstrap the LLM’s responses on argument filling prompts, so that its generative behavior can be controlled to output the arguments in a dictionary format, as illustrated in~\figurename~\ref{fig:prompt template}.
Following the conventional fine-tuning scheme, we fine-tune the LLM using the cross entropy loss.
% For both STAR and SGD datasets, we fine-tune the LLM for 5 epochs, 
% We observe that after 5 epochs of bootstrapping, the LLM starts to output fairly well-formatted arguments.
% for reward function calculation.
Once the bootstrapping phase is completed, we propose to augment the train dataset using model-generated outputs.
In the next section, we define a custom reward function that is employed to score and select generated samples to be included in the additional fine-tuning phase. \\ 
% and identify arguments that will be incorporated in the additional fine-tuning phase to further improve the argument filling performance. \\ 
% ; only the arguments that surpass a pre-set threshold are used to augment the train dataset for the additional fine-tuning phase. \\

\noindent\textbf{Phase II. Rejection Sampling with Custom Reward Function}
"Rejection Sampling" commonly refers to the process of identifying desirable model-generated outputs that are capable of further improving the performance on the target task.
Therefore, the success of rejection sampling is heavily contingent on the definition of the reward function that can accurately reflect the usefulness of model-generated outputs. 
To define the custom reward function for argument filling, we first categorize potential sources of error into: non-existent key (NK), missing key (MK), schema-grounded but incorrect value (SV), and hallucinated value (HV).
The key and value here refer to the corresponding components of the key-value pairs of the model-generated arguments, which have been bootstrapped to follow a dictionary-like format.
A detailed description of each error type is provided below:\\
\noindent$\bullet$~~\textbf{Non-existent Key (NK):} The generated key is not provided as a part of the pre-defined schema.  \\
\noindent$\bullet$~~\textbf{Missing Key (MK):} The model-generated arguments are missing an expected key that is required by the pre-defined schema. \\
\noindent$\bullet$~~\textbf{Schema-grounded but Incorrect Value (SV):} The generated value follows the pre-defined schema but deviates from the dialogue history, resulting in an incorrectly identified argument. \\
\noindent$\bullet$~~\textbf{Hallucinated Value (HV):} The generated value does not follow the pre-defined schema, and hence, it is incorrect by definition. \\
% Detailed description of each error type is provided in Section~\ref{error_type} of Appendix.
The total number of errors in a model-generated output can be computed through a simple summation of all 4 error types: $N_\mathrm{Error} = N_\mathrm{NK} + N_\mathrm{MK} + N_\mathrm{SV} + N_\mathrm{HV} $.
The error rate can then be defined as: $N_\mathrm{Error} / N_\mathrm{Total},$ where $N_\mathrm{Total}$ denotes the total number of keys and values in the ground-truth argument. 
This error rate is normalized between $-1$ and $1$ to obtain the final reward value following the equation: $R = 1 - 2 * N_\mathrm{Error} / N_\mathrm{Total}$.

After the LLM has been bootstrapped on the argument filling datasets, we sample $K$ number of outputs from the model and score the generated outputs using the above reward function. 
We only select outputs that yield positive reward to augment the train dataset.
With the newly added instances mixed in the train dataset, we perform one additional epoch of supervised fine-tuning.

There exist two expected advantages of incorporating rejection-sampled model outputs.
First, utilizing the model outputs filtered with the custom reward function allows us to effectively augment the train dataset with desirable instances without the need to collect additional data points to avoid overfitting. 
Second, we expect that incorporating these outputs will improve the fine-tuned LLM’s robustness to noisy data points it may encounter at test-time. 
Even if the model-generated outputs yield positive reward, they will inevitably be noisier than the curated train dataset with ground-truth labels. 
Therefore, the LLM that has been exposed to noisier data points in the rejection sampling phase will exhibit a higher degree of robustness and generalization performance. 

\subsection{Multi-step Prompting Scheme for Closed-sourced LLMs}~\label{multistep}
It is infeasible to fine-tune LLMs whose design and weights are not released to the public.
% It may be infeasible to fine-tune larger LLMs in a resource-constrained setting.
Therefore, we additionally explore a more fine-grained and informative prompting method to complement larger LLMs. 
The default prompt design as described in~\figurename~\ref{fig:prompt template} asks the model to identify required arguments and extract appropriate information to fill them all at the same time. 
For multi-step prompting with hints, we instead prompt the model to identify and fill one argument at a time. 
By using this more targeted prompt design, we are providing the LLM with additional information about required slots and effectively restricting its generative behavior to prevent its digression from the pre-defined schema and dialogue history.
% From here on, we will refer to this prompt design as ''multi-turn prompting."

\section{Experimental Set-up}
\subsection{Datasets and Models}
\subsubsection{Datasets}
We primarily use STAR~\cite{mosig2020star} and SGD~\cite{rastogi2020towards} datasets as test beds to validate our approach. \\
\noindent$\bullet$~~\textbf{STAR:} is a collection of realistic, task-oriented dialogues that includes 5,820 dialogues that span 24 tasks and 13 domains. The schemas in the STAR dataset are similar to "task specifications,” which contain information about the ideal dialogue flow for each task. \\
\noindent$\bullet$~~\textbf{SGD:} is a rich, fully-annotated dataset, which contains more than 22,000 dialogues that encompass 20 domains, ranging from banks to travels and weather. The comprehensive annotation that includes schema representation makes it a flexible and convenient dataset to investigate not only argument filling but also other components of task-oriented conversational systems. \\
We verify the competitiveness of proposed approaches under both in- and out-of-domain scenarios. 
Under the in-domain scenario, train and test dialogues are sampled from the same set of domains, while under the out-of-domain scenario, the test dialogues contain domains that were not observed during the training process.
To create an in-domain benchmark, we randomly split the entire dataset into train and test datasets, such that domains are evenly represented across the two.
For out-of-domain evaluation, we purposefully curate the test dataset, such that no explicit or semantic overlap exist between tasks in the train dataset and those in the test dataset. 
% are guaranteed to lie outside the train task distribution.

\begin{table*}[t]
\centering
% \vspace{-5pt}
\setlength{\tabcolsep}{6pt}
% \vspace{-3pt}
\renewcommand{\arraystretch}{1.0}
\begin{tabular}{l|l|c|c|c|c|c|c}
\toprule
\multirow{2}{*}{Methods} & \multirow{2}{*}{Models} & \multicolumn{3}{c|}{SGD} & \multicolumn{3}{c}{STAR} \\
& & BLEU & FM & F-1 & BLEU & FM & F-1 \\
\bottomrule
\toprule
\multirow{2}{*}{Zero-shot}  & LLAMA-v1-7B & 0.0104 & 5.2852 & 0.0472 & \multicolumn{2}{c}{------}  \\
% & LLAMA-v2-13B-Chat & 0.0191 & 13.3109 & 0.0879 & 
%  0.0496 & 19.1845 & 0.0918 \\
& ChatGPT & 0.4578 & 44.5853 & 0.4802 & 0.2127 & 26.0679 & 0.2094 \\
\midrule
Multi-step & \multirow{2}{*}{ChatGPT} & \multirow{2}{*}{0.4578} & \multirow{2}{*}{44.5853} & \multirow{2}{*}{0.4802} & \multirow{2}{*}{0.2127} & \multirow{2}{*}{26.0679} & \multirow{2}{*}{0.2094} \\
Prompting & & & & & & & \\
\midrule
% Multi-step & LLAMA-v2-13B-Chat & 0.0775 & 38.6579 & 0.2908 & 0.1233 & 21.8156 & 0.1774 \\
% Prompting & ChatGPT & 0.5073 & 60.98 & 0.5892 &  0.1427 & 32.7379 & 0.2019 \\
% \midrule
\multirow{2}{*}{Instruction-tuned} & LLAMA-v1-7B-\textit{sft} & 0.7802 & 91.1299 & 0.7718 & 0.3418 & 58.19 & 0.3209 \\
& LLAMA-v1-7B-\textit{sft-rs} & 0.8003 & 91.6462 & 0.7834 & 0.3734 & 62.7669 & 0.3605 \\
\bottomrule
\end{tabular}
\vspace{-5pt}
\caption{Comparison of different models and training/prompting methods under the \textbf{in-domain} evaluation setting. LLAMA-v1-7B\textit{-sft-rs} clearly outperforms all other baselines, showing the efficacy of the proposed training scheme.}
\vspace{-10pt}
\label{table:main_indomain}
\end{table*}

\begin{table*}[t]
\centering
% \vspace{-5pt}
\setlength{\tabcolsep}{6pt}
% \vspace{-3pt}
\renewcommand{\arraystretch}{1.0}
\begin{tabular}{l|l|c|c|c|c|c|c}
\toprule
\multirow{2}{*}{Methods} & \multirow{2}{*}{Models} & \multicolumn{3}{c|}{SGD} & \multicolumn{3}{c}{STAR} \\
& & BLEU & FM & F-1 & BLEU & FM & F-1 \\
\bottomrule
\toprule
\multirow{2}{*}{Zero-shot}  & LLAMA-v1-7B & 0.0118 & 5.4612 & 0.0456 & \multicolumn{3}{c}{------}  \\
% & LLAMA-v2-13B-Chat & 0.0179 & 14.5424 & 0.0643 & 0.0231 & 14.73 & 0.0529 \\
& ChatGPT & 0.2460 & 35.9156 & 0.3701 & 0.2045 & 33.5571 & 0.2357 \\
\midrule
% Multi-turn & LLAMA-v2-13B-Chat & 0.0349 & 25.9211 & 0.1366 & 0.1113 & 20.0857 & 0.1727 \\
Multi-step & \multirow{2}{*}{ChatGPT} & \multirow{2}{*}{0.3166} & \multirow{2}{*}{48.7200} & \multirow{2}{*}{0.4212} & \multirow{2}{*}{0.2281} & \multirow{2}{*}{33.7857} & \multirow{2}{*}{0.2672} \\
Prompting & & & & & & & \\
\midrule
\multirow{2}{*}{Instruction-tuned} & LLAMA-v1-7B-\textit{sft} & 0.6972 & 86.3976 & 0.6642 & 0.2512 & 54.7000 & 0.2330 \\
& LLAMA-v1-7B-\textit{sft-rs} & 0.7705 & 90.6652 & 0.7608 & 0.3511 & 65.0714 & 0.3200 \\
\bottomrule
\end{tabular}
\vspace{-5pt}
\caption{Comparison of different models and training/prompting methods under the \textbf{out-of-domain} evaluation setting. The results are generally consistent with those obtained under the in-domain setting.}
\vspace{-10pt}
\label{table:main_outofdomain}
\end{table*}

\begin{figure}[t]
\begin{center}
  \includegraphics[width=\linewidth]{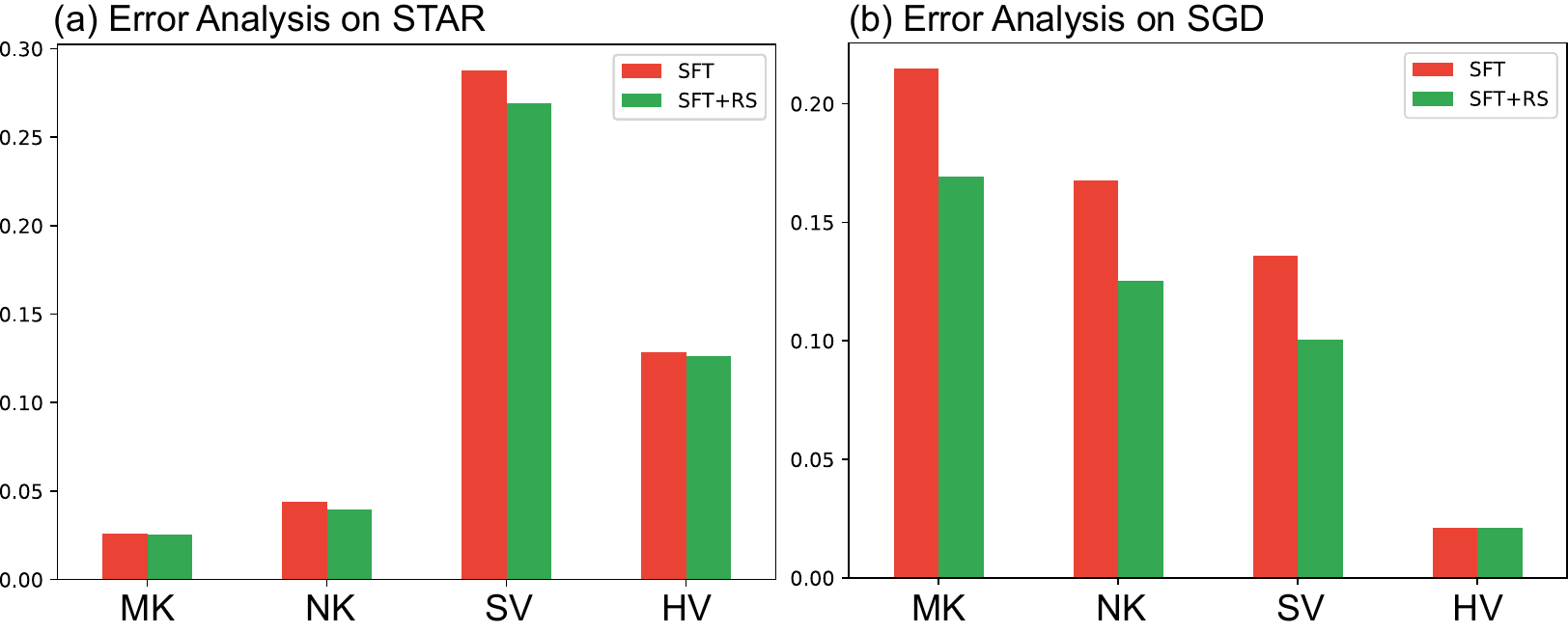}
\end{center}
  \caption{Analyses of four different error rates on (a) STAR and (b) SGD \textbf{in-domain} benchmarks.}
\vspace{-5pt}
\label{fig:ind_error}
\end{figure}

\subsubsection{Models}
\noindent$\bullet$~~\textbf{LLAMA-7B~\cite{touvron2023llama}:} is a state-of-the-art foundational LLM released by Meta AI. While the LLAMA models of various sizes have been open-sourced, we primarily utilize LLAMA-v1-7B model for fine-tuning experiments. \\
% \noindent$\bullet$~~\textbf{LLAMAv2-13B-Chat:} is a more recent, advanced version of the LLAMA model that has been trained with 40\% more data. We utilize the instruction-tuned chat version of the LLAMA-v2-13B model to construct a strong open-source baseline. \\
\noindent$\bullet$~~\textbf{ChatGPT\footnote{\url{https://openai.com/blog/chatgpt}}:} is widely regarded as one of the most powerful LLMs; its release is perceived to be a significant milestone in the evolution of conversational AI systems. Because the model weights have not been open-sourced, we rely on OpenAI’s ChatGPT API for evaluation.

\subsection{Libraries and Hyperparameters}
We utilize the Huggingface~\cite{wolf2019huggingface} library for implementation and training of models.
All experiments are executed on NVIDIA V100 GPU with 32GB RAM.
The following set of hyperparameters is used for the supervised fine-tuning phase: batch size of 8, Adam optimizer with initial learning rate of 0.00002, weight decay of 0.1, and constant learning rate scheduling.
We run the supervised fine-tuning phase for 5 epochs before performing rejection sampling.
As mentioned in Section~\ref{inst_tuning}, we perform additional fine-tuning with rejection-sampled data for only one additional epoch.
All hyperparameters remain unchanged from the supervised fine-tuning phase.

\subsection{Compared Approaches}
\noindent$\bullet$~~\textbf{Zero-shot}: is the most na\"ve baseline obtained by prompting the pre-trained LLMs with the prompt design provided in~\figurename~\ref{fig:prompt template}. The pre-trained LLMs are used as is without undergoing additional fine-tuning on task-oriented dialogue datasets. \\
\noindent$\bullet$~~\textbf{Multi-Step}: replaces the na\"ve prompting process with the multi-step prompting scheme in Section~\ref{multistep}. Since multi-step prompting only improves the model at inference time, the pre-trained LLM is again used with no alterations. \\
\noindent$\bullet$~~\textbf{Supervised Fine-tuning}~(-\textit{sft}): is a baseline obtained by instruction-tuning the LLM on fully-labeled train set of task-oriented dialogue datasets following the Phase I process in Section~\ref{inst_tuning}. \\
% The input to the model is formatted following the prompt design in Section 3.(a), and the required arguments in the form of dictionary (e.g., {Name: Angela, Location:  South Lake Union, Menu: Coffee}) are used as groundtruth labels. We use the plain cross-entropy loss to for supervised fine-tuning. \\
\noindent$\bullet$~~\textbf{Supervised Fine-tuning + Our Rejection Sampling}~(-\textit{sft}-\textit{rs}): trains the fine-tuned LLM on additional model-generated data that have been selected according to the proposed reward for rejection sampling (Phase II of Section~\ref{inst_tuning}).
%Please refer to Section 3.(b) for the detailed description of the proposed framework. 

\subsection{Metrics}
\noindent$\bullet$~~\textbf{BLEU:}~\cite{papineni2002bleu} quantifies the semantic similarity between model-generated and reference sentence pairs. 
Its close alignment with human perception of generation quality and low computational cost make BLEU a particularly compelling metric for automatic evaluation of Natural Language Processing (NLP) systems. \\
% is one of the most commonly-used metrics in Natural Language Processing (NLP) and is representative of the
\noindent$\bullet$~~\textbf{Fuzzy Matching:} is adopted to quantify the argument filling accuracy. We employ fuzzy match, instead of exact match, such that minor typos and capitalization, which should not determine the quality of the generated outputs, do not influence the performance metric. \\ 
% (how much the model-generated outputs match the ground-truth arguments) is computed as the number of dialogue turns for which the the predicted argument matches the ground truth divided by the total number of dialogue turns. To determine if the predicted argument is a match, we will be using fuzzy matching, such that minor typos and capitalization, which do not affect the quality of the generative response, do not influence the performance metric. \\
\noindent$\bullet$~~\textbf{F-1 Score:} takes into account both the character-level precision and recall of predicted arguments. F-1 score is a preferred choice of metric over accuracy when evaluating datasets with significant class imbalances (i.e., the number of test samples per API is unevenly distributed).

\begin{figure}[t]
\begin{center}
  \includegraphics[width=\linewidth]{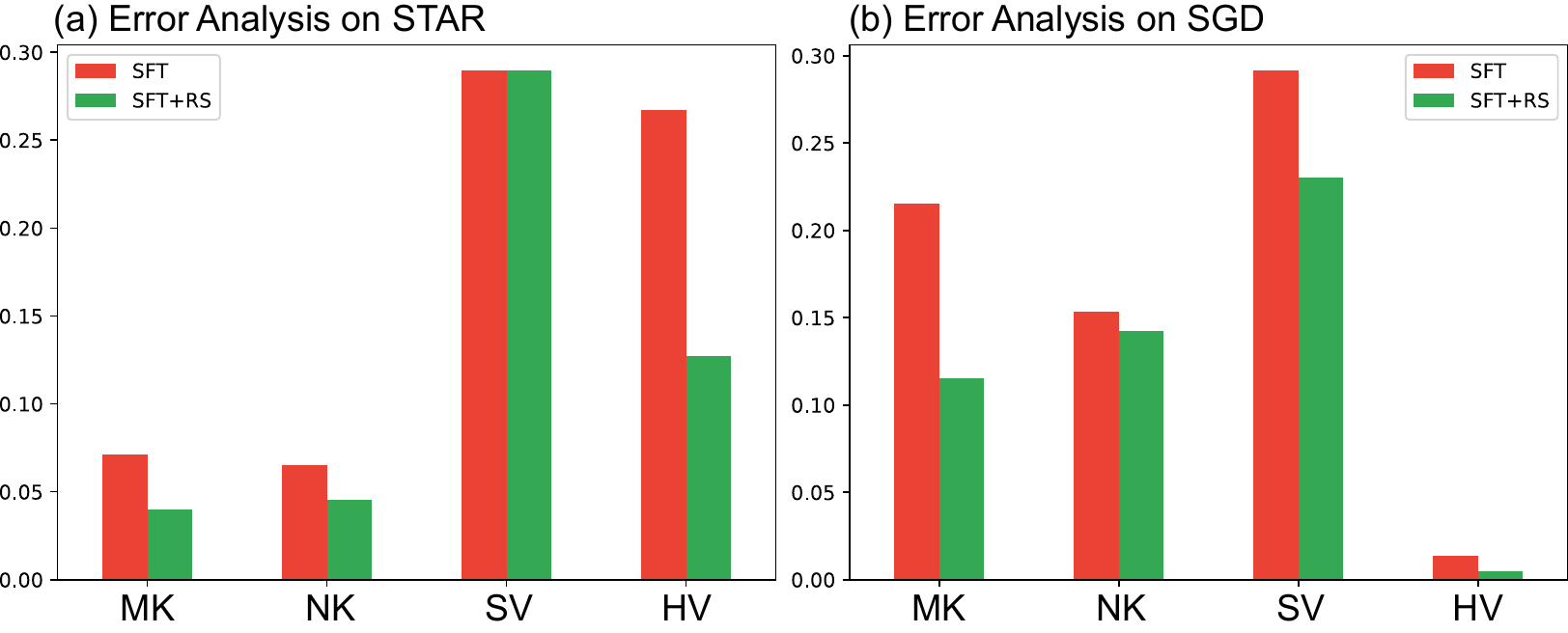}
\end{center}
  \caption{Analyses of four different error rates on (a) STAR and (b) SGD \textbf{out-of-domain} benchmarks.}
\vspace{-5pt}
\label{fig:ood_error}
\end{figure}

\section{Results}
\subsection{In-Domain Results}
The results obtained on STAR and SGD datasets under the in-domain evaluation setting are reported in~\tablename~\ref{table:main_indomain}.
The suffixes \textit{-sft} and \textit{-sft-rs} are used to denote models that have been trained only with supervised fine-tuning and with supervised fine-tuning and rejection sampling, respectively.
Multi-step prompting that provides additional hints successfully improves the performance of the ChatGPT models.
More importantly, we observe that the LLAMA-v1-7B model that has been trained with the proposed instruction-tuning pipeline with rejection sampling (LLAMA-v1-7B\textit{-sft-rs}) obtains the best performance across all metrics on both datasets.
This result clearly demonstrates that with our training framework, relatively smaller and light-weight LLMs can outperform larger ones.
Furthermore, the superiority of LLAMA-v1-7B\textit{-sft-rs} to LLAMA-v1\textit{-sft} provides strong support for incorporating rejection-sampled data to effectively improve the performance of fine-tuning with less training budget.
Lastly, we note that a larger degree of performance improvement is observed on the SGD dataset, which has a wider variety of tasks and thus can be considered more difficult.
% LLAMA-v1 fine-tuned with the SFT + RS pipeline comes on top on both datasets across all metrics.
% With our SFT + RS pipeline, relatively smaller and light-weight LLM, like LLAMA-v1, can outperform larger ones.
% Such a result showcases the effectiveness of the proposed instruction-tuning pipeline in argument filling.

\subsection{Out-of-Domain Results}
To simulate an out-of-domain test scenario, we deliberately create a train-test split, such that there is no explicit or implicit task domain overlap between the train and test set
The results obtained under the out-of-domain evaluation setting are reported in~\tablename~\ref{table:main_outofdomain}. 
In general, the out-of-domain evaluation results show similar tendencies to the in-domain results.
While the proposed instruction-tuning framework and multi-step prompting successfully improve the performance of open-sourced and closed-sourced LLMs, respectively, they both experience slight performance degradation when compared to the in-domain evaluation results.
% To evaluate the performance of compared models on out-of-domain tasks, we deliberately create a train-test split, such that there is no explicit or implicit task domain overlap between the train and test set

\subsection{Error Analyses}
We analyze sources of error in outputs generated by LLAMA-v1-7B\textit{-sft-rs} to identify room for improvement.
In Figures~\ref{fig:ind_error} and~\ref{fig:ood_error}, we compare the four error rates, as defined in Section~\ref{inst_tuning}, in LLAMA-v1-7B\textit{-sft} and LLAMA-v1-7B\textit{-sft-rs} models.
Training the LLAMA-v1-7B model with SFT $+$ RS reduces all four error rates, and the rate of hallucinated value errors is particularly low compared to other errors.
This analytical result implies that once grounded, the LLM mostly ceases to hallucinate and remains close to the API schema and dialogue history provided as a part of the prompt template.

\section{Conclusion}
This paper explored and uncovered the powerfulness of leveraging LLMs to automate the argument filling process, a core component in task-oriented conversational systems.
The strong experimental results indicate that the proposed methods, used in conjunction with open- or closed-source LLMs, are effective for restricting the LLM's generative behavior, specifically for argument filling. 

\section*{Acknowledgements}
This work was supported in part by the Institute of Information \& Communications Technology Planning \& Evaluation (IITP) and the National Research Foundation of Korea (NRF) grants funded by the Korean government (MSIT) [No. 2021-0-01343, No. 2022-0-00959, Artificial Intelligence Graduate School Program (Seoul National University), No. 2022R1A3B1077720] and the BK21 FOUR program of the Education and Research Program for Future ICT Pioneers, Seoul National University in 2024.
% BK: This work was supported by .
% AI: This work was supported by Institute of Information & Communications Technology Planning & Evaluation (IITP) grant funded by the Korea government (MSIT) [NO.2021-0-01343, Artificial Intelligence Graduate School Program (Seoul National University)
% 2022R1A3B1077720 -> 리더연구

\section*{Limitations and Potential Risks}
One limitation of our work is that proposed frameworks are validated only on one open- or closed-sourced model.
In addition, while LLMs are quite capable of completing the argument filling task, the inference time for LLMs may still be longer than many of smaller, more targeted language models.
Accelerating LLM inferencing, however, is outside the scope of our work.

Reliance on closed-sourced LLMs could pose unforeseen risks since the backbone model could be altered without notice.
Even if significant changes are made to the design and weights of the closed-sourced models, there is no way for us to know what those alterations are.
This complete black-box nature of closed-sourced LLMs may make it an undesirable choice of backbone model.
Therefore, we conjecture that utilizing a targeted decoding scheme that can further enforce the LLM to follow specific parts of the prompt template could assist in reducing schema-related errors.

% We believe that  to a greater variety of models would be necessary to demonstrate that the findings reported here are not specific to the studied models.

% \clearpage 

% \section{Ethical Considerations}
% \section{Conclusion}
% \clearpage

\bibliography{main}

\begin{thebibliography}{26}
\expandafter\ifx\csname natexlab\endcsname\relax\def\natexlab#1{#1}\fi

\bibitem[{Brown et~al.(2020)Brown, Mann, Ryder, Subbiah, Kaplan, Dhariwal,
  Neelakantan, Shyam, Sastry, Askell et~al.}]{brown2020language}
Tom Brown, Benjamin Mann, Nick Ryder, Melanie Subbiah, Jared~D Kaplan, Prafulla
  Dhariwal, Arvind Neelakantan, Pranav Shyam, Girish Sastry, Amanda Askell,
  et~al. 2020.
\newblock Language models are few-shot learners.
\newblock \emph{Advances in neural information processing systems},
  33:1877--1901.

\bibitem[{Cai et~al.(2019)Cai, Wang, Bi, Tu, Liu, and Shi}]{cai2019retrieval}
Deng Cai, Yan Wang, Wei Bi, Zhaopeng Tu, Xiaojiang Liu, and Shuming Shi. 2019.
\newblock Retrieval-guided dialogue response generation via a
  matching-to-generation framework.
\newblock In \emph{Proceedings of the 2019 Conference on Empirical Methods in
  Natural Language Processing and the 9th International Joint Conference on
  Natural Language Processing (EMNLP-IJCNLP)}, pages 1866--1875.

\bibitem[{Goldwasser and Roth(2014)}]{goldwasser2014learning}
Dan Goldwasser and Dan Roth. 2014.
\newblock Learning from natural instructions.
\newblock \emph{Machine learning}, 94(2):205--232.

\bibitem[{Hosseini-Asl et~al.(2020)Hosseini-Asl, McCann, Wu, Yavuz, and
  Socher}]{hosseini2020simple}
Ehsan Hosseini-Asl, Bryan McCann, Chien-Sheng Wu, Semih Yavuz, and Richard
  Socher. 2020.
\newblock A simple language model for task-oriented dialogue.
\newblock \emph{Advances in Neural Information Processing Systems},
  33:20179--20191.

\bibitem[{Hude{\v{c}}ek and Du{\v{s}}ek(2023)}]{hudevcek2023llms}
Vojt{\v{e}}ch Hude{\v{c}}ek and Ond{\v{r}}ej Du{\v{s}}ek. 2023.
\newblock Are llms all you need for task-oriented dialogue?
\newblock \emph{arXiv preprint arXiv:2304.06556}.

\bibitem[{Kulh{\'a}nek et~al.(2021)Kulh{\'a}nek, Hude{\v{c}}ek, Nekvinda, and
  Du{\v{s}}ek}]{kulhanek2021augpt}
Jon{\'a}{\v{s}} Kulh{\'a}nek, Vojt{\v{e}}ch Hude{\v{c}}ek, Tom{\'a}{\v{s}}
  Nekvinda, and Ond{\v{r}}ej Du{\v{s}}ek. 2021.
\newblock Augpt: Auxiliary tasks and data augmentation for end-to-end dialogue
  with pre-trained language models.
\newblock In \emph{Proceedings of the 3rd Workshop on Natural Language
  Processing for Conversational AI}, pages 198--210.

\bibitem[{Lin et~al.(2020)Lin, Madotto, Winata, and Fung}]{lin2020mintl}
Zhaojiang Lin, Andrea Madotto, Genta~Indra Winata, and Pascale Fung. 2020.
\newblock Mintl: Minimalist transfer learning for task-oriented dialogue
  systems.
\newblock In \emph{Proceedings of the 2020 Conference on Empirical Methods in
  Natural Language Processing (EMNLP)}, pages 3391--3405.

\bibitem[{Liu and Low(2023)}]{liu2023goat}
Tiedong Liu and Bryan Kian~Hsiang Low. 2023.
\newblock Goat: Fine-tuned llama outperforms gpt-4 on arithmetic tasks.
\newblock \emph{arXiv preprint arXiv:2305.14201}.

\bibitem[{Mishra et~al.(2022)Mishra, Khashabi, Baral, and Hajishirzi}]{NIv1}
Swaroop Mishra, Daniel Khashabi, Chitta Baral, and Hannaneh Hajishirzi. 2022.
\newblock Cross-task generalization via natural language crowdsourcing
  instructions.
\newblock In \emph{Proceedings of the 60th Annual Meeting of the Association
  for Computational Linguistics (Volume 1: Long Papers)}, pages 3470--3487.

\bibitem[{Mosig et~al.(2020)Mosig, Mehri, and Kober}]{mosig2020star}
Johannes~EM Mosig, Shikib Mehri, and Thomas Kober. 2020.
\newblock Star: A schema-guided dialog dataset for transfer learning.
\newblock \emph{arXiv preprint arXiv:2010.11853}.

\bibitem[{Nekvinda and Du{\v{s}}ek(2022)}]{nekvinda2022aargh}
Tom{\'a}{\v{s}} Nekvinda and Ond{\v{r}}ej Du{\v{s}}ek. 2022.
\newblock Aargh! end-to-end retrieval-generation for task-oriented dialog.
\newblock \emph{arXiv preprint arXiv:2209.03632}.

\bibitem[{Ouyang et~al.(2022)Ouyang, Wu, Jiang, Almeida, Wainwright, Mishkin,
  Zhang, Agarwal, Slama, Ray et~al.}]{ouyang2022training}
Long Ouyang, Jeffrey Wu, Xu~Jiang, Diogo Almeida, Carroll Wainwright, Pamela
  Mishkin, Chong Zhang, Sandhini Agarwal, Katarina Slama, Alex Ray, et~al.
  2022.
\newblock Training language models to follow instructions with human feedback.
\newblock \emph{Advances in Neural Information Processing Systems},
  35:27730--27744.

\bibitem[{Pandey et~al.(2018)Pandey, Contractor, Kumar, and
  Joshi}]{pandey2018exemplar}
Gaurav Pandey, Danish Contractor, Vineet Kumar, and Sachindra Joshi. 2018.
\newblock Exemplar encoder-decoder for neural conversation generation.
\newblock In \emph{Proceedings of the 56th Annual Meeting of the Association
  for Computational Linguistics (Volume 1: Long Papers)}, pages 1329--1338.

\bibitem[{Papineni et~al.(2002)Papineni, Roukos, Ward, and
  Zhu}]{papineni2002bleu}
Kishore Papineni, Salim Roukos, Todd Ward, and Wei-Jing Zhu. 2002.
\newblock Bleu: a method for automatic evaluation of machine translation.
\newblock In \emph{Proceedings of the 40th annual meeting of the Association
  for Computational Linguistics}, pages 311--318.

\bibitem[{Peng et~al.(2021)Peng, Li, Li, Shayandeh, Liden, and
  Gao}]{peng2021soloist}
Baolin Peng, Chunyuan Li, Jinchao Li, Shahin Shayandeh, Lars Liden, and
  Jianfeng Gao. 2021.
\newblock Soloist: Building task bots at scale with transfer learning and
  machine teaching.
\newblock \emph{Transactions of the Association for Computational Linguistics},
  9:807--824.

\bibitem[{Raheja et~al.(2023)Raheja, Kumar, Koo, and Kang}]{raheja2023coedit}
Vipul Raheja, Dhruv Kumar, Ryan Koo, and Dongyeop Kang. 2023.
\newblock Coedit: Text editing by task-specific instruction tuning.
\newblock \emph{arXiv preprint arXiv:2305.09857}.

\bibitem[{Rastogi et~al.(2020)Rastogi, Zang, Sunkara, Gupta, and
  Khaitan}]{rastogi2020towards}
Abhinav Rastogi, Xiaoxue Zang, Srinivas Sunkara, Raghav Gupta, and Pranav
  Khaitan. 2020.
\newblock Towards scalable multi-domain conversational agents: The
  schema-guided dialogue dataset.
\newblock In \emph{Proceedings of the AAAI Conference on Artificial
  Intelligence}, volume~34, pages 8689--8696.

\bibitem[{Touvron et~al.(2023)Touvron, Lavril, Izacard, Martinet, Lachaux,
  Lacroix, Rozi{\`e}re, Goyal, Hambro, Azhar et~al.}]{touvron2023llama}
Hugo Touvron, Thibaut Lavril, Gautier Izacard, Xavier Martinet, Marie-Anne
  Lachaux, Timoth{\'e}e Lacroix, Baptiste Rozi{\`e}re, Naman Goyal, Eric
  Hambro, Faisal Azhar, et~al. 2023.
\newblock Llama: Open and efficient foundation language models.
\newblock \emph{arXiv preprint arXiv:2302.13971}.

\bibitem[{Vaswani et~al.(2017)Vaswani, Shazeer, Parmar, Uszkoreit, Jones,
  Gomez, Kaiser, and Polosukhin}]{vaswani2017attention}
Ashish Vaswani, Noam Shazeer, Niki Parmar, Jakob Uszkoreit, Llion Jones,
  Aidan~N Gomez, {\L}ukasz Kaiser, and Illia Polosukhin. 2017.
\newblock Attention is all you need.
\newblock \emph{Advances in neural information processing systems}, 30.

\bibitem[{Wang et~al.(2023)Wang, Zhou, Zu, Xia, Chen, Zhang, Zheng, Ye, Zhang,
  Gui et~al.}]{wang2023instructuie}
Xiao Wang, Weikang Zhou, Can Zu, Han Xia, Tianze Chen, Yuansen Zhang, Rui
  Zheng, Junjie Ye, Qi~Zhang, Tao Gui, et~al. 2023.
\newblock Instructuie: Multi-task instruction tuning for unified information
  extraction.
\newblock \emph{arXiv preprint arXiv:2304.08085}.

\bibitem[{Wang et~al.(2022{\natexlab{a}})Wang, Mishra, Alipoormolabashi, Kordi,
  Mirzaei, Arunkumar, Ashok, Dhanasekaran, Naik, Stap et~al.}]{NIv2}
Yizhong Wang, Swaroop Mishra, Pegah Alipoormolabashi, Yeganeh Kordi, Amirreza
  Mirzaei, Anjana Arunkumar, Arjun Ashok, Arut~Selvan Dhanasekaran, Atharva
  Naik, David Stap, et~al. 2022{\natexlab{a}}.
\newblock Benchmarking generalization via in-context instructions on 1,600+
  language tasks.
\newblock \emph{arXiv preprint arXiv:2204.07705}.

\bibitem[{Wang et~al.(2022{\natexlab{b}})Wang, Mishra, Alipoormolabashi, Kordi,
  Mirzaei, Naik, Ashok, Dhanasekaran, Arunkumar, Stap et~al.}]{wang2022super}
Yizhong Wang, Swaroop Mishra, Pegah Alipoormolabashi, Yeganeh Kordi, Amirreza
  Mirzaei, Atharva Naik, Arjun Ashok, Arut~Selvan Dhanasekaran, Anjana
  Arunkumar, David Stap, et~al. 2022{\natexlab{b}}.
\newblock Super-naturalinstructions: Generalization via declarative
  instructions on 1600+ nlp tasks.
\newblock In \emph{Proceedings of the 2022 Conference on Empirical Methods in
  Natural Language Processing}, pages 5085--5109.

\bibitem[{Wolf et~al.(2019)Wolf, Debut, Sanh, Chaumond, Delangue, Moi, Cistac,
  Rault, Louf, Funtowicz et~al.}]{wolf2019huggingface}
Thomas Wolf, Lysandre Debut, Victor Sanh, Julien Chaumond, Clement Delangue,
  Anthony Moi, Pierric Cistac, Tim Rault, R{\'e}mi Louf, Morgan Funtowicz,
  et~al. 2019.
\newblock Huggingface's transformers: State-of-the-art natural language
  processing.
\newblock \emph{arXiv preprint arXiv:1910.03771}.

\bibitem[{Yunxiang et~al.(2023)Yunxiang, Zihan, Kai, Ruilong, and
  You}]{yunxiang2023chatdoctor}
Li~Yunxiang, Li~Zihan, Zhang Kai, Dan Ruilong, and Zhang You. 2023.
\newblock Chatdoctor: A medical chat model fine-tuned on llama model using
  medical domain knowledge.
\newblock \emph{arXiv preprint arXiv:2303.14070}.

\bibitem[{Zhang et~al.(2022)Zhang, Roller, Goyal, Artetxe, Chen, Chen, Dewan,
  Diab, Li, Lin et~al.}]{zhang2022opt}
Susan Zhang, Stephen Roller, Naman Goyal, Mikel Artetxe, Moya Chen, Shuohui
  Chen, Christopher Dewan, Mona Diab, Xian Li, Xi~Victoria Lin, et~al. 2022.
\newblock Opt: Open pre-trained transformer language models.
\newblock \emph{arXiv preprint arXiv:2205.01068}.

\bibitem[{Zhang et~al.(2019)Zhang, Sun, Galley, Chen, Brockett, Gao, Gao, Liu,
  and Dolan}]{zhang2019dialogpt}
Yizhe Zhang, Siqi Sun, Michel Galley, Yen-Chun Chen, Chris Brockett, Xiang Gao,
  Jianfeng Gao, Jingjing Liu, and Bill Dolan. 2019.
\newblock Dialogpt: Large-scale generative pre-training for conversational
  response generation.
\newblock \emph{arXiv preprint arXiv:1911.00536}.

\end{thebibliography}
\clearpage
\appendix

% \section{Appendix}
% \label{sec:appendix}

% \setcounter{section}{0}
% \renewcommand\thesection{A\arabic{section}}
% \setcounter{table}{0}
% \renewcommand{\thetable}{A\arabic{table}}
% \setcounter{figure}{0}
% \renewcommand{\thefigure}{A\arabic{figure}}
% \setcounter{equation}{0}
% \renewcommand{\theequation}{A\arabic{equation}}

% \section{Error Types for Rejection Sampling}~\label{error_type}

% \section{Experimental Details}~\label{exp_det}

\end{document}